%% file: ijcai26.tex
\documentclass{article}
\usepackage{ijcai26}

\usepackage{times}
\usepackage{soul}
\usepackage{url}
\usepackage[hidelinks]{hyperref}
\usepackage[utf8]{inputenc}
\usepackage[small]{caption}
\usepackage{graphicx}
\usepackage{amsmath}
\usepackage{amssymb}
\usepackage{amsthm}
\usepackage{booktabs}
\usepackage{algorithm}
\usepackage{algorithmic}
\usepackage[switch]{lineno}

\title{ManiSplat: Manipulation Trajectory Synthesis from Monocular Video via Decoupled 3D Gaussian Splatting}

\author{
Wenhao Hu$^{1,2}$\footnote{This work was done during an internship at Horizon Robotics.} \and
Haonan Zhou$^1$\and
Liu Liu$^2$\footnote{Project lead.}\and
Yun Du$^2$\and
Xinjie Wang$^2$\and
Ziang Li$^2$\and 
Zhizhong Su$^2$\And
Gaoang Wang$^1$\footnote{Corresponding author.} \\
\affiliations
$^1$Zhejiang University\\
$^2$Horizon Robotics\\
}

\begin{document}

\maketitle

\input{sec/0_abs}

\input{sec/1_intro}
\input{sec/2_related}

\input{sec/3_method}

\input{sec/4_exp}

\input{sec/5_conclusion}

\section*{Acknowledgments}
This work was supported by the National Natural Science Foundation of China (No. 62576308), Zhejiang Provincial Natural Science Foundation of China (No. LZ24F030005), and Fundamental Research Funds for the Central Universities (No. 226-2025-00167).
\bibliographystyle{named}
\bibliography{ijcai26}

\end{document}

%% file: sec/0_abs.tex
\begin{abstract}
Reconstructing dynamic and interactive 3D scenes from real-world observations remains a fundamental challenge in computer vision and robotics. While recent advances in 3D Gaussian Splatting have enabled high-fidelity static reconstruction, extending it to interactive environments with articulated robots and manipulable objects remains difficult due to complex contact interactions and abrupt pose changes. To address these challenges, we introduce ManiSplat, a unified framework that reconstructs controllable and decoupled Gaussian digital twins directly from monocular ego-view robotic videos. Our method introduces a Graph-Structured Disentangled Representation that separates the robot, objects, and background into independently optimizable Gaussian subfields organized within a scene graph. To ensure stability, we propose a Task-Oriented Spatio-Temporal Alignment module that leverages the inherent logic of manipulation tasks—alternating between Motion and Skill phases—to construct accurate pseudo-ground-truth trajectories. Finally, a joint photometric-geometric optimization ensures the reconstructed scenes are temporally coherent, physically consistent, and simulation-ready. Extensive experiments demonstrate that our approach reconstructs interaction-driven dynamic scenes with high fidelity and controllability, effectively supporting downstream robotic tasks and policy learning. The project page is available at \url{https://whhu7.github.io/ManiSplat/}.
\end{abstract}

%% file: sec/1_intro.tex
\section{Introduction}
\label{sec:intro}
Reconstructing dynamic and interactive 3D scenes from real-world observations remains a fundamental challenge in computer vision and robotics. Most existing dynamic-Gaussian frameworks treat Gaussians as functions of time, implicitly encoding motion within the representation. Methods such as Deformable 3D Gaussian \cite{yang2024deformable}, 4D-GS \cite{wu20244d}, MotionGS \cite{zhu2024motiongs}, and Spacetime Gaussian \cite{li2024spacetime} enable high-quality dynamic rendering but lack explicit object-level disentanglement. Later works, including Ex4DGS \cite{lee2024fully}, HUGS \cite{zhou2024hugs}, EgoGaussian \cite{zhang2025egogaussian}, SplitGaussian \cite{li2025splitgaussian}, and BézierGS \cite{Ma2025BezierGS}, explicitly separate static and dynamic components to improve geometric stability and temporal coherence.
However, these approaches primarily target non-interactive dynamic scenes—such as driving or human motion—where dynamics arise naturally and are only observed rather than induced.
They remain limited in interaction-driven scenarios, where articulated agents (e.g., robotic arms) actively manipulate objects, producing complex contact interactions and pose changes. Although IGFuse~\cite{hu2026igfuse} explores interactive Gaussian reconstruction by fusing multi-state scene scans, it relies on discrete observations with object rearrangements rather than continuous robotic manipulation videos.

\begin{figure}[t]
  \centering
  \includegraphics[width=1.0\linewidth]{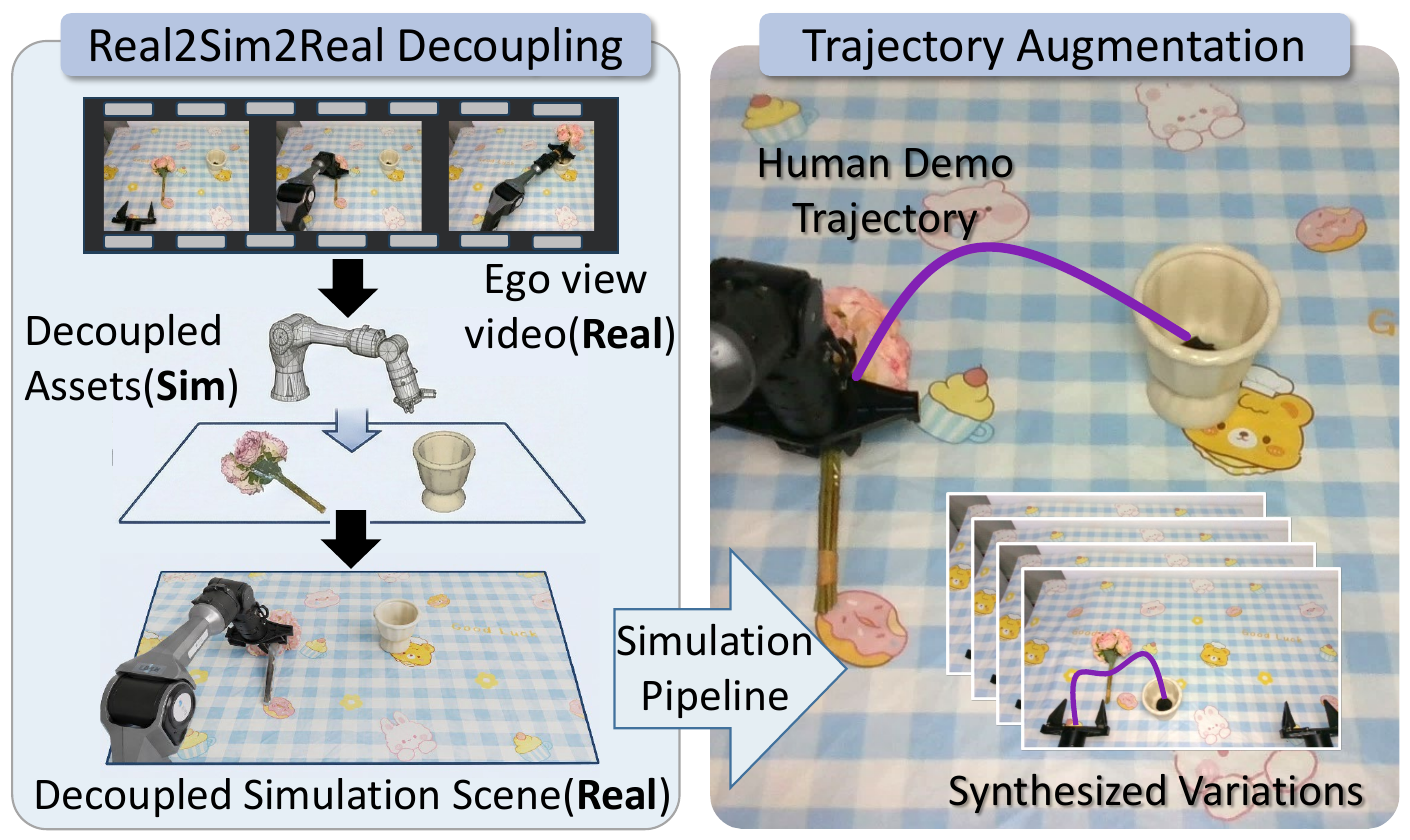}
 \caption{\textbf{High-Fidelity Real2Sim Alignment and Trajectory Augmentation.} 
  \textbf{(Left)} Our framework decomposes monocular ego-view video into aligned assets (URDF robot, objects, and background). 
  \textbf{(Right)} This enables a trajectory augmentation pipeline that synthesizes diverse variations from single demonstrations to scale up data for policy learning.}
  
  \label{fig:teaser}
\end{figure}

Meanwhile, the Real2Sim2Real (R2S2R) paradigm has emerged as an effective strategy to bridge real-world perception and simulation, enabling consistent transfer of sensing and control for robotic learning\cite{8793789}.
Recent studies have incorporated 3DGS into robotic pipelines to enhance photorealism and physical consistency.
For instance, Robo-GS \cite{lou2025robo}, RL-GSBridge \cite{wu2025rl}, SplatSim \cite{qureshi2025splatsim}, and RoboGSim \cite{li2024robogsim} couple Gaussian Splatting with physics-aware constraints to achieve differentiable rendering and physically consistent control;
DREMA \cite{barcellona2024dream} and RE3SIM \cite{han2025re} further construct learnable digital twins for embodied policy learning.
However, most of these frameworks rely on static scene scans and pre-aligned assets, making it difficult to model robots directly from dynamically captured videos.

To address these limitations, we propose a unified framework for reconstructing interaction-driven dynamic Gaussian scenes directly from monocular ego-view robotic videos. First, we design a Graph-Structured Disentangled Representation that explicitly separates the robot arm, manipulable objects, and static background into independently optimizable Gaussian subfields within a scene graph. Second, we present a Task-Oriented Spatio-Temporal Alignment module. This module leverages the inherent logic of manipulation tasks—alternating between Motion and Skill phases—to construct accurate pseudo-ground-truth pose trajectories through Hybrid Pose Estimation and appearance alignment, bridging the domain gap while ensuring stability.

Finally, leveraging this decoupled and controllable representation, we develop a Topology-Preserving Data Augmentation pipeline. By adopting a ``skill-preserving, motion-adaptive" strategy, we can apply rigid transformations to identified skill segments while autonomously re-planning motion segments. This mechanism enables the generation of large-scale, physically feasible, and spatially diverse synthetic trajectories from a single real-world demonstration, providing high-fidelity data support for downstream embodied policy learning.

In summary, our main contributions are as follows:
\begin{itemize}
    \item We propose a unified framework for reconstructing interactive and simulation-ready Gaussian scenes from real-world monocular ego-view robotic videos.
    \item We design a graph-based disentangled representation system that achieves independent modeling and object-level controlled reconstruction of the robot, objects, and background.
    \item We introduce a task-oriented spatio-temporal alignment mechanism that significantly improves trajectory stability and rendering fidelity during interactions through hybrid pose estimation and appearance constraints.
    \item We develop a topology-preserving data augmentation method capable of synthesizing physically consistent augmented trajectories from a single demonstration, effectively addressing data sparsity and enhancing policy generalization.
\end{itemize}

%% file: sec/2_related.tex
\section{Related Works}
\label{sec:related}
\subsection{Real2Sim2Real}
The Real2Sim2Real (R2S2R) paradigm bridges real-world data and simulation for robotic learning, enabling consistent transfer of perception and control.
Recent efforts integrate 3D Gaussian Splatting (3DGS) into robotic pipelines to enhance photorealism and physical realism.
Robo-GS~\cite{lou2025robo} fuses mesh geometry, Gaussian kernels, and physics attributes via a Gaussian–Mesh–Pixel binding mechanism for differentiable, physically grounded rendering.
RL-GSBridge~\cite{wu2025rl} and SplatSim~\cite{qureshi2025splatsim} embed 3DGS within reinforcement learning frameworks, ensuring visually consistent control and reducing Sim2Real gaps.
RoboGSim~\cite{li2024robogsim}, DREMA~\cite{barcellona2024dream}, and RE3SIM~\cite{han2025re} further integrate Gaussian splatting with physics simulation to construct learnable, photorealistic digital twins for policy learning.
RoboSplat~\cite{yang2025novel} directly edits reconstructed Gaussian scenes to generate diverse robotic demonstrations.
Unlike these methods that depend on static scene scans for geometry and pose alignment, our approach performs video-based dynamic disentanglement and fusion, leveraging temporal motion cues from real videos to unify appearance and pose across the real and simulated domains.

\subsection{Dynamic Gaussian Reconstruction}
Recent advances in dynamic scene modeling extend static 3D Gaussian splatting to the temporal domain.
Deformable 3D Gaussian~\cite{yang2024deformable} reconstructs dynamic scenes in a canonical space via a learned deformation field, enabling real-time, temporally smooth rendering.
4D-GS~\cite{wu20244d} jointly encodes space and time through 3D Gaussians coupled with 4D neural voxels for compact, high-resolution rendering.
MotionGS~\cite{zhu2024motiongs} introduces optical-flow-based motion priors to decouple camera and object motion, improving monocular reconstruction accuracy.
Spacetime Gaussian~\cite{li2024spacetime} augments 3D Gaussians with temporal opacity and motion attributes to achieve high-fidelity, view- and time-dependent rendering.
MEGA~\cite{zhang2025mega} and FreeTimeGS~\cite{wang2025freetimegs} further improve efficiency and flexibility through compact color encoding and learnable motion functions, respectively.
However, these approaches generally treat Gaussians as temporal functions and fail to disentangle individual moving objects from the scene.
To address this, recent works such as Ex4DGS~\cite{lee2024fully}, HUGS~\cite{zhou2024hugs}, EgoGaussian~\cite{zhang2025egogaussian}, SplitGaussian~\cite{li2025splitgaussian}, and BézierGS~\cite{Ma2025BezierGS} explicitly separate static and dynamic components, enabling more stable and consistent motion reconstruction.
Nevertheless, existing methods are primarily designed for passive dynamic scenes (e.g., driving or human motion) and cannot effectively model robotic manipulation scenarios, where object trajectories are actively induced by articulated agents.
In contrast, our framework explicitly models interaction-driven dynamics within Gaussian fields, bridging the gap between dynamic reconstruction and physically grounded manipulation understanding.

%% file: sec/3_method.tex
\section{Preliminary}
\subsection{Gaussian Splatting}
Following ~\cite{kerbl20233d}, a 3D scene is represented as a set of Gaussian primitives 
$\mathcal{G} = \{\mathbf{x},\, \boldsymbol{\Sigma},\, \alpha,\, \mathbf{c}\}$,
where $\mathbf{x}$ denotes the 3D center position, $\boldsymbol{\Sigma}$ is the spatial covariance matrix,
$\alpha$ is the opacity coefficient, and $\mathbf{c}$ is the RGB color vector. 
During rendering, each Gaussian is projected onto the 2D image plane through a differentiable $\alpha$-blending process, and the final pixel color $\mathbf{C}$ is computed by accumulating Gaussian contributions along the viewing ray as 
$\mathbf{C} = \sum_{i \in \mathcal{N}} \mathbf{c}_i \alpha_i' \prod_{j<i} (1 - \alpha_j')$, 
where $\mathcal{N}$ denotes the ordered set of Gaussians intersected by the ray.  

\begin{figure*}[t]
  \centering
  \includegraphics[width=1.0\linewidth]{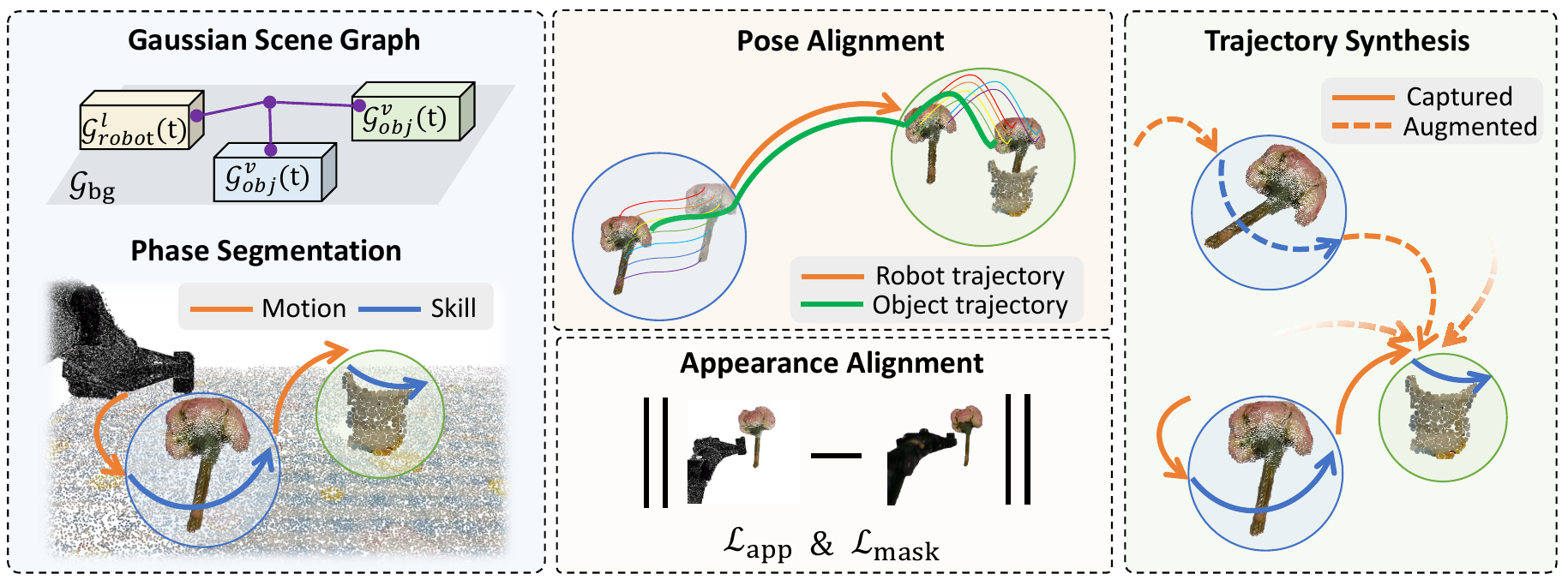}
  \caption{
  \textbf{Overview of the Proposed Framework.} 
  \textbf{(Left)} We construct a \textit{Gaussian Scene Graph} ($\mathcal{G}$) to disentangle the scene into independent semantic nodes: the robot $\mathcal{G}_{robot}$, the object $\mathcal{G}_{obj}$, and the static background $\mathcal{G}_{bg}$. Simultaneously, the manipulation task is segmented into a robot-centric \textit{Motion} phase (orange, e.g., transport) and an object-centric \textit{Skill} phase (blue, e.g., insertion).
  \textbf{(Middle)} To build a high-fidelity digital twin, we perform joint optimization via \textit{Pose Alignment} and \textit{Appearance Alignment}.
  \textbf{(Right)} Leveraging the decoupled structure, we perform \textit{Trajectory Synthesis} by generating diverse augmented approach paths (dashed orange lines) that seamlessly converge into the preserved skill execution, significantly scaling up the demonstration data.
  }
  \label{fig:method}
\end{figure*}

We further define the application of a rigid transformation $T = (R, t) \in \mathrm{SE}(3)$ to a Gaussian as $T \otimes \mathcal{G} = \{R\mathbf{x} + t,\, R\boldsymbol{\Sigma}R^{\top},\, \alpha,\, \mathbf{c}\}$, 
where $R$ and $t$ denote the rotation and translation components, respectively. 
This operation transforms the Gaussian’s center and covariance into world coordinates while keeping its opacity and color unchanged.

\section{Method}
\label{sec:method}

As illustrated in Fig.~\ref{fig:method}, our framework reconstructs controllable Gaussian scenes from real robotic videos and generates diverse, physically feasible demonstration data. The system consists of three core components: \textbf{Graph-Structured Disentangled Representation} (\S\ref{sec:graph}) models the robot, objects, and background as independent Gaussian subfields to achieve a structured scene representation. \textbf{Task-Oriented Spatio-Temporal Alignment} (\S\ref{sec:align}) leverages interaction logic (Motion vs. Skill) to construct pseudo-ground-truth object poses, while performing appearance alignment to ensure the rendered Gaussians are consistent with real-world observations in both pose and visual features. \textbf{Topology-Preserving Data Augmentation} (\S\ref{sec:aug}) synthesizes new trajectory data by rigidly transforming skills and re-planning motions, significantly scaling up the demonstration data. Finally, we present our unified optimization objective in \S\ref{sec:opt}.

\subsection{Graph-Structured Disentangled Representation}
\label{sec:graph}

\paragraph{Gaussian Scene Graph.}
To strictly disentangle the scene for downstream control, we represent the world as a dynamic scene graph comprising three fundamental types of nodes:
(1) The Background Node ($G_{\text{bg}}$), which contains static Gaussians representing the environment;
(2) The Robot Node ($G_{\text{robot}}$), modeled as articulated Gaussian clusters driven by the URDF kinematic chain;
(3) The Object Nodes ($G_{\text{obj}}^{v}$), representing manipulable tools or rigid entities.

At any time step $t$, the complete scene is composed via the union of these transformed subfields:
\begin{equation}
    G(t) = G_{\text{bg}} \cup G_{\text{robot}}(t) \cup \{G_{\text{obj}}^{v}(t)\}_v
\end{equation}

\paragraph{Robot Node.}
The robot geometry is explicitly driven by its kinematic state. Given joint angles $\mathbf{q}_t \in \mathbb{R}^{n}$, the Gaussian primitives attached to link $l$, denoted as $\bar{G}_{\text{robot}}^{(l)}$, are transformed to world space via forward kinematics $T_{\text{base}\to l}(\mathbf{q}_t)$:
\begin{equation}
    G_{\text{robot}}^{(l)}(t) = T_{\text{base}\to l}(\mathbf{q}_t) \otimes \bar{G}_{\text{robot}}^{(l)},
\end{equation}
where $\otimes$ denotes the application of an $\mathrm{SE}(3)$ transformation to the Gaussian centers and covariances.

\paragraph{Object Nodes.}
Each object node $v$ maintains a canonical Gaussian set $\bar{G}_{\text{obj}}^{v}$ and a time-varying pose $T_{v}(t) \in \mathrm{SE}(3)$. The world-space representation is computed as:
\begin{equation}
    G_{\text{obj}}^{v}(t) = T_{v}(t) \otimes \bar{G}_{\text{obj}}^{v}.
\end{equation}
By decoupling object-specific Gaussians from the background, our framework enables high-fidelity rendering of object interactions and supports flexible trajectory manipulation.

\subsection{Task-Oriented Spatio-Temporal Alignment}
\label{sec:align}

Accurate estimation of object pose $T_{v}(t)$ is critical for reconstruction. Instead of relying on noisy optical flow or heuristic smoothing, we propose a task-oriented approach that utilizes the inherent structure of manipulation tasks—alternating between free-space motion and contact-rich interaction.

\paragraph{Interaction-Aware Phase Segmentation.} 
Following the semantic decomposition in DemoGen~\cite{xue2025demogen}, we partition the demonstration into two distinct phases based on the proximity between the robot end-effector (EE) and the target object. Let $d(t)$ denote the distance between the EE and the object center. We define a proximity threshold $\tau$ to classify the trajectory: the \textbf{Skill Phase ($S$)} occurs when $d(t) < \tau$, representing critical interactions like grasping or placing; otherwise, the segment is classified as the \textbf{Motion Phase ($M$)}, representing free-space transport or approaching. For a typical pick-and-place task, the sequence is naturally segmented as: $M_{\text{approach}} \to S_{\text{pick}} \to M_{\text{transfer}} \to S_{\text{place}} \to M_{\text{return}}$.

\paragraph{Hybrid Pose Estimation.}
Based on these phases, we construct a reliable pseudo-ground-truth pose trajectory, denoted as $T_{v}^{\text{pseudo}}(t)$, by applying phase-specific geometric constraints:

\begin{enumerate}
    \item \textbf{Static Constraint ($M_{\text{approach}}, M_{\text{return}}$):}
    When the object is not involved in manipulation, it remains static in the world frame. We fix its pose to the initial or final stable state::
    \begin{equation}
        T_{v}^{\text{pseudo}}(t) =
        \begin{cases}
            T_{v}(t_{\text{init}}), & \forall t \in M_{\text{approach}},\\
            T_{v}(t_{\text{final}}), & \forall t \in M_{\text{return}}.
        \end{cases}
    \end{equation}
    where $T_{v}(t_{\text{init}})$ denotes the initial stable pose of the object before the task starts, and $T_{v}(t_{\text{final}})$ denotes the final stable pose of the object after the task is completed.

    \item \textbf{Attached Constraint ($M_{\text{transfer}}$):} 
    During transport, the object is rigidly attached to the gripper. We assume the relative transformation $T_{\text{rel}}$ (captured at the end of the grasp skill) remains constant. The object pose is derived purely from robot kinematics:
    \begin{equation}
        T_{v}^{\text{pseudo}}(t) = T_{\text{ee}}(t) \cdot T_{\text{rel}}, \quad \forall t \in M_{\text{transfer}}
    \end{equation}
    where $T_{\text{ee}}(t)$ is the accurate end-effector pose from the robot controller.

    \item \textbf{Tracking-Based Constraint ($S_{\text{pick}}, S_{\text{place}}$):} 
    During interaction, the object may undergo complex movements that are neither static nor fully rigid with the EE. To capture this, we employ a point tracker to track a set of 3D keypoints $\mathcal{P} = \{\mathbf{p}_k\}$ on the object. We solve for the optimal pose $T(t)$ that minimizes the registration error:
    \begin{equation}
        T_{v}^{\text{pseudo}}(t) = \mathop{\arg\min}_{T \in \mathrm{SE}(3)} \sum_{k} \| T \cdot \mathbf{p}_k(0) - \mathbf{p}_k(t) \|^2
    \end{equation}
\end{enumerate}

\paragraph{Appearance Alignment.}
Lighting and texture differences between the URDF-rendered robot arm and its real counterpart often lead to a significant appearance gap. To bridge this domain gap, we utilize SAM2~\cite{ravi2024sam} to generate dynamic masks $M_{\text{dyn}}$ for the moving components. During training, appearance consistency is enforced via masked $\mathcal{L}_{1}$ and $\mathcal{L}_{\text{SSIM}}$ losses:
\begin{equation}
    \mathcal{L}_{\text{app}} = \lambda_{1} \mathcal{L}_{1}(I \cdot M_{\text{dyn}}, \hat{I} \cdot M_{\text{dyn}}) + \lambda_{\text{ssim}} \mathcal{L}_{\text{SSIM}}(I \cdot M_{\text{dyn}}, \hat{I} \cdot M_{\text{dyn}}).
\end{equation}
To further disentangle dynamic objects from the static background, we impose a mask-consistency constraint between the rendered alpha map of dynamic Gaussians $\mathcal{O}^{G}_{\text{dyn}}$ and the SAM2 mask $M_{\text{dyn}}$:
\begin{equation}
    \mathcal{L}_{\text{mask}} = \|\mathcal{O}^{G}_{\text{dyn}} - M_{\text{dyn}}\|_{1}.
\end{equation}
Enforcing these alignment losses not only bridges the visual gap but also promotes coarse pose alignment, as visual consistency implicitly guides the geometric optimization of the scene graph.

\subsection{Optimization}
\label{sec:opt}

\begin{figure*}[!t]
    \centering
    \includegraphics[width=1\linewidth]{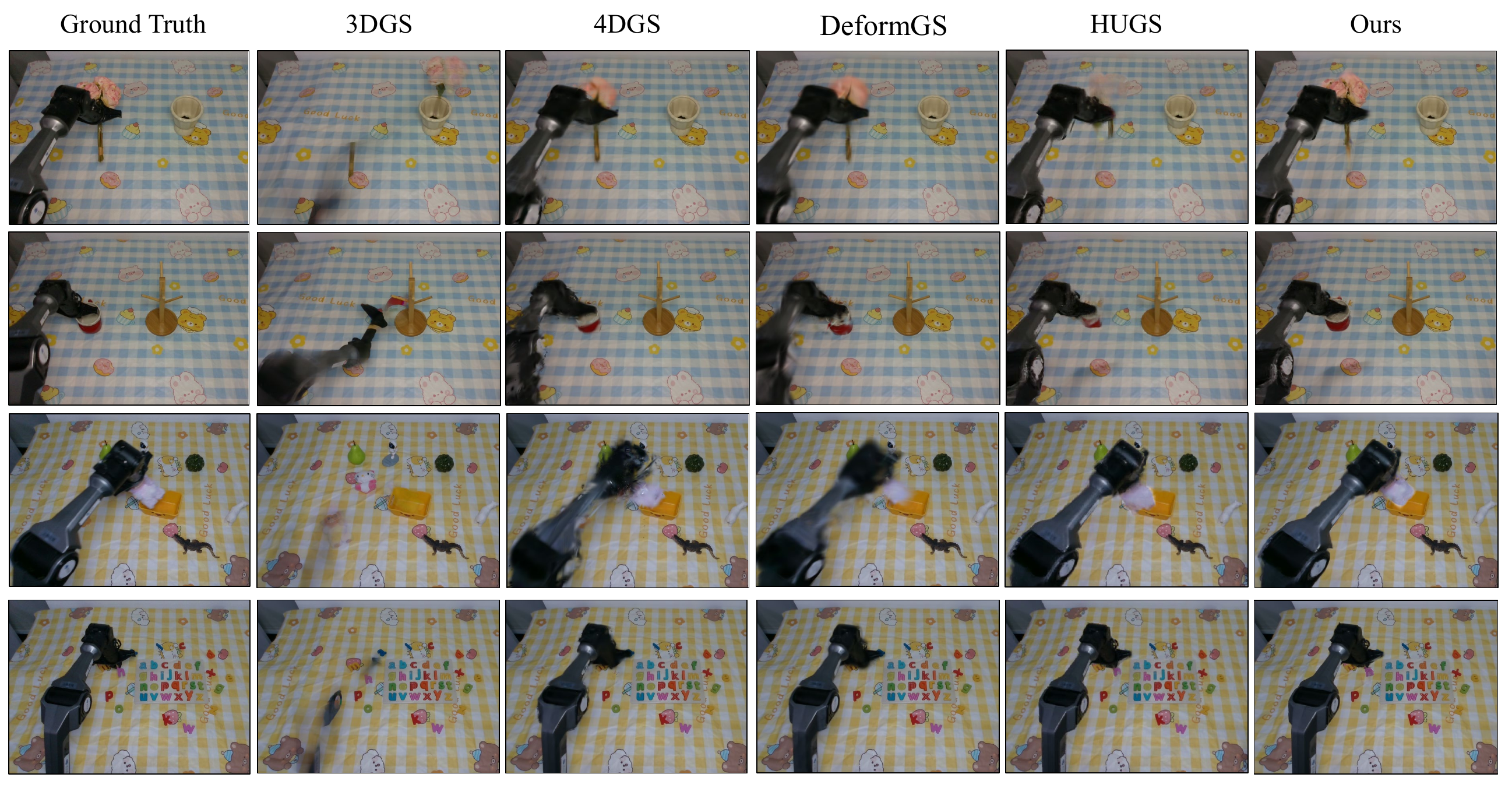}
    \caption{Qualitative comparison of dynamic reconstruction. 4DGS and DeformableGS exhibit artifacts and blurring during rapid manipulator movements due to the lack of decoupling. HUGS produces blurry object details by over-smoothing the abrupt pose changes during grasping. In contrast, our method maintains sharp boundaries and stable appearance by explicitly modeling the interaction stages.}
    \label{fig:dynamic_qualitative}
\end{figure*}

We jointly optimize the Gaussian parameters and the scene graph structure. In addition to the standard photometric loss used in vanilla 3D Gaussian Splatting~\cite{kerbl20233d}, which ensures pixel-level color and structural consistency, we incorporate strong geometric and appearance constraints to handle the challenges of robotic manipulation scenes.

Specifically, we introduce a pose loss $\mathcal{L}_{\text{pose}}$ to supervise the object trajectories based on our hybrid estimation:
\begin{equation}
    \mathcal{L}_{\text{pose}} = \sum_{t} \| T_{v}(t) \ominus T_{v}^{\text{pseudo}}(t) \|_1,
\end{equation}
where $\ominus$ denotes the distance metric in $\mathrm{SE}(3)$, covering both rotation and translation.

The complete optimization objective is formulated as a weighted sum of the rendering, appearance, mask, and pose terms:
\begin{equation}
    \mathcal{L}_{\text{total}} = \lambda_{\text{render}}\mathcal{L}_{\text{render}} + \lambda_{\text{app}}\mathcal{L}_{\text{app}} + \lambda_{\text{mask}}\mathcal{L}_{\text{mask}} + \lambda_{\text{pose}}\mathcal{L}_{\text{pose}}.
\end{equation}
By integrating these losses, our framework ensures that the reconstructed scene is not only visually high-fidelity but also physically consistent with the robot's kinematic constraints, facilitating effective Real2Sim2Real transfer.

\subsection{Topology-Preserving Data Augmentation}
\label{sec:aug}

To scale up demonstration data for policy learning, we generate synthetic trajectories by modifying the scene layout while preserving the underlying task logic. We adopt a ``Skill-Preserving, Motion-Adaptive'' strategy that augments object poses and ensures kinematic feasibility through autonomous re-planning.

\paragraph{Skill Augmentation via Rigid Transformation.}
Manipulation skills rely on the local relative pose between the gripper and the object. For an identified Skill segment $S_i$, we apply a random rigid transformation $T_{\text{aug}} \in \mathrm{SE}(3)$ to the object's pose. To maintain the integrity of the interaction mechanics, the end-effector trajectory is subjected to the same transformation:
\begin{equation}
    T_{\text{ee}}'(t) = T_{\text{aug}} \cdot T_{\text{ee}}(t), \quad \forall t \in S_i
\end{equation}
This operation ensures that delicate actions, such as grasping or insertion, remain geometrically invariant and are simply transposed to a new task coordinates defined by $T_{\text{aug}}$.

\paragraph{Motion Generation via Re-planning.}
Motion segments serve as transitions between consecutive skills. Directly transforming the original motion trajectory often results in collisions or kinematic singularities within the augmented layout. To address this, we treat each Motion segment $M_i$ as a navigation problem between the augmented skills. Let the terminal state of the preceding augmented skill be $\mathbf{q}_{\text{start}}'$ and the initial state of the subsequent augmented skill be $\mathbf{q}_{\text{goal}}'$. We invoke a motion planning framework to generate a new, collision-free trajectory that connects these states. This decoupled approach enables the robot to adaptively navigate the modified environment while strictly adhering to the core manipulation logic.

%% file: sec/4_exp.tex
\section{Experiments}
\label{sec:exp}
\subsection{Dataset}
\label{dataset}
To comprehensively evaluate our method, we utilize both synthetic and real-world datasets. The synthetic data is generated using the RoboTwin simulation platform~\cite{chen2025robotwin}, which provides high-fidelity rendering and precise ground truth parameters. The real-world data is captured using an AgileX dual-arm manipulator, recording complex interaction sequences in physical environments.

For the reconstruction task, we employ a frame-subsampling strategy to partition the data into training and testing sets, evaluating the novel view synthesis quality. For the pose estimation task, we rely on the synthetic ground truth pose trajectories to quantitatively assess the accuracy of our estimations.

\subsection{Experimental Setup}

\paragraph{Implementation details.}
The training process is conducted for a total of 30,000 iterations using the Adam optimizer. To ensure robust scene initialization, we obtain the initial object poses via GenPose++~\cite{zhang2024omni6dpose}. For the segments requiring dynamic tracking, CoTracker~\cite{karaev2025cotracker3} is employed to provide point-based estimation. The distance threshold $\tau$, which distinguishes between the motion and skill phases, is set to 0.2. The loss balancing hyperparameters are assigned as $\lambda_{\text{app}} = 1.0$, $\lambda_{\text{mask}} = 1.0$, and $\lambda_{\text{pose}} = 1.0$. All experiments and runtime evaluations are performed on a single NVIDIA RTX 4090 GPU.

\paragraph{Baselines.} We compare our proposed method against several representative Gaussian Splatting-based frameworks, categorized by their modeling capabilities. First, we include \textbf{3DGS}~\cite{kerbl20233d} as a baseline for static scene reconstruction, serving as a reference that lacks dynamic modeling capabilities. To evaluate dynamic scene modeling, we compare against \textbf{4DGS}~\cite{wu20244d} and \textbf{DeformableGS}~\cite{yang2024deformable}. While these methods effectively handle non-rigid deformations, they do not explicitly decouple the object from the environment. Finally, we examine decoupled dynamic reconstruction methods, including \textbf{HUGS}~\cite{zhou2024hugs}, which employs function fitting for pose curves. To ensure a fair comparison with the latter, we extend the dynamics function within HUGS to support full 3D spatial modeling.

\subsection{Dynamic Reconstruction}
\label{dynamic_recon}

\begin{table*}[t]
    \centering
    \resizebox{\linewidth}{!}{
        \begin{tabular}{l c c cc cc cc}
            \toprule
            \cmidrule(lr){4-9}
            & & & \multicolumn{2}{c}{Full Image} & \multicolumn{2}{c}{Robot} & \multicolumn{2}{c}{Object} \\
            \cmidrule(lr){4-5} \cmidrule(lr){6-7} \cmidrule(lr){8-9}
            Methods & \textit{Dynamic} & \textit{Decouple} & PSNR$\uparrow$ & SSIM$\uparrow$ & PSNR$\uparrow$ & SSIM$\uparrow$ & PSNR$\uparrow$ & SSIM$\uparrow$ \\
            \midrule
            3DGS \cite{kerbl20233d}  & &  & 19.93 & 0.898 & 5.36 & 0.415 & 12.57 & 0.633 \\
            4DGS\cite{wu20244d}& \checkmark&  & 30.68 & 0.937 & 21.37 & 0.824 & 23.41 & 0.865 \\
            DeformGS\cite{yang2024deformable}&\checkmark &  & 30.24 & 0.940 & 20.20 & 0.818 & 22.96 & 0.863 \\
            HUGS\cite{zhou2024hugs} & \checkmark & \checkmark & 30.70 & 0.893 & 26.07& 0.838 & 18.90 & 0.767 \\
            \midrule
            \textbf{Ours} & \checkmark & \checkmark & \textbf{32.31} & \textbf{0.951} & \textbf{26.07} & \textbf{0.838} & \textbf{26.85} & \textbf{0.872} \\
            \bottomrule
        \end{tabular}
    }
    \caption{Quantitative comparison of scene reconstruction. We compare our method with state-of-the-art methods on full image, robot, and object regions.}
    \label{tab:dynamic_metrics}
\end{table*}

\paragraph{Qualitative Analysis.}
The qualitative performance of dynamic reconstruction is illustrated in Fig.~\ref{fig:dynamic_qualitative}, while quantitative results are detailed in Tab.~\ref{tab:dynamic_metrics}. Due to the lack of explicit object-environment decoupling, 4DGS and DeformableGS struggle to handle the rapid motion of the robotic arm, often failing to deform correctly and resulting in significant artifacts and motion blur. While decoupled methods address this to some extent, they face distinct challenges in pose estimation. HUGS employs dynamics curve fitting to constrain motion; however, during high-degree-of-freedom interactions—such as the moment of grasping—the strict smoothing function tends to obliterate the necessary abrupt changes in object pose, similarly leading to blurred reconstruction. Our method addresses these limitations through a multi-stage pose handling strategy. In the motion phase, we leverage robot kinematics to maintain relative static attributes, while in the abrupt skill phase, we incorporate tracking supervision. This allows us to maintain stability in both pose and appearance throughout the entire interaction sequence.

\paragraph{Quantitative Analysis.}
Quantitatively, 3DGS yields the lowest PSNR as it lacks dynamic modeling capabilities. While 4DGS and DeformableGS improve upon this with dynamic reconstruction, their inability to decouple the scene results in poor performance when evaluating the manipulator and object separately. For fair comparison, we supply HUGS with ground-truth robot joint angles, resulting in manipulator reconstruction quality comparable to ours. However, their object reconstruction metrics are significantly lower, as HUGS suffers from over-smoothing in the presence of complex contact. Our method achieves superior performance across all metrics by accurately capturing the complex interaction dynamics.

\subsection{Pose Estimation}
\label{pose_est}

\begin{table}[!t]
    \centering
    \begin{tabular}{lcc}
        \toprule
        Method & $E_t$ (cm) $\downarrow$ & $E_R$ ($^\circ$) $\downarrow$ \\
        \midrule
        GenPose++~\cite{zhang2024omni6dpose} & 0.9965 & 7.109 \\
        HUGS~\cite{zhou2024hugs} & 0.9936 & 7.004 \\
        \midrule
        \textbf{Ours} & \textbf{0.5864} & \textbf{5.185} \\
        \bottomrule
    \end{tabular} 
    \caption{\textbf{Quantitative comparison on RoboTwin.} We report the average translation error ($E_t$) in cm and rotation error ($E_R$) in degrees. The best results are highlighted in \textbf{bold}.}
    \label{tab:pose_error}
\end{table}

\begin{figure}[!t]
    \centering
    \includegraphics[width=1\linewidth]{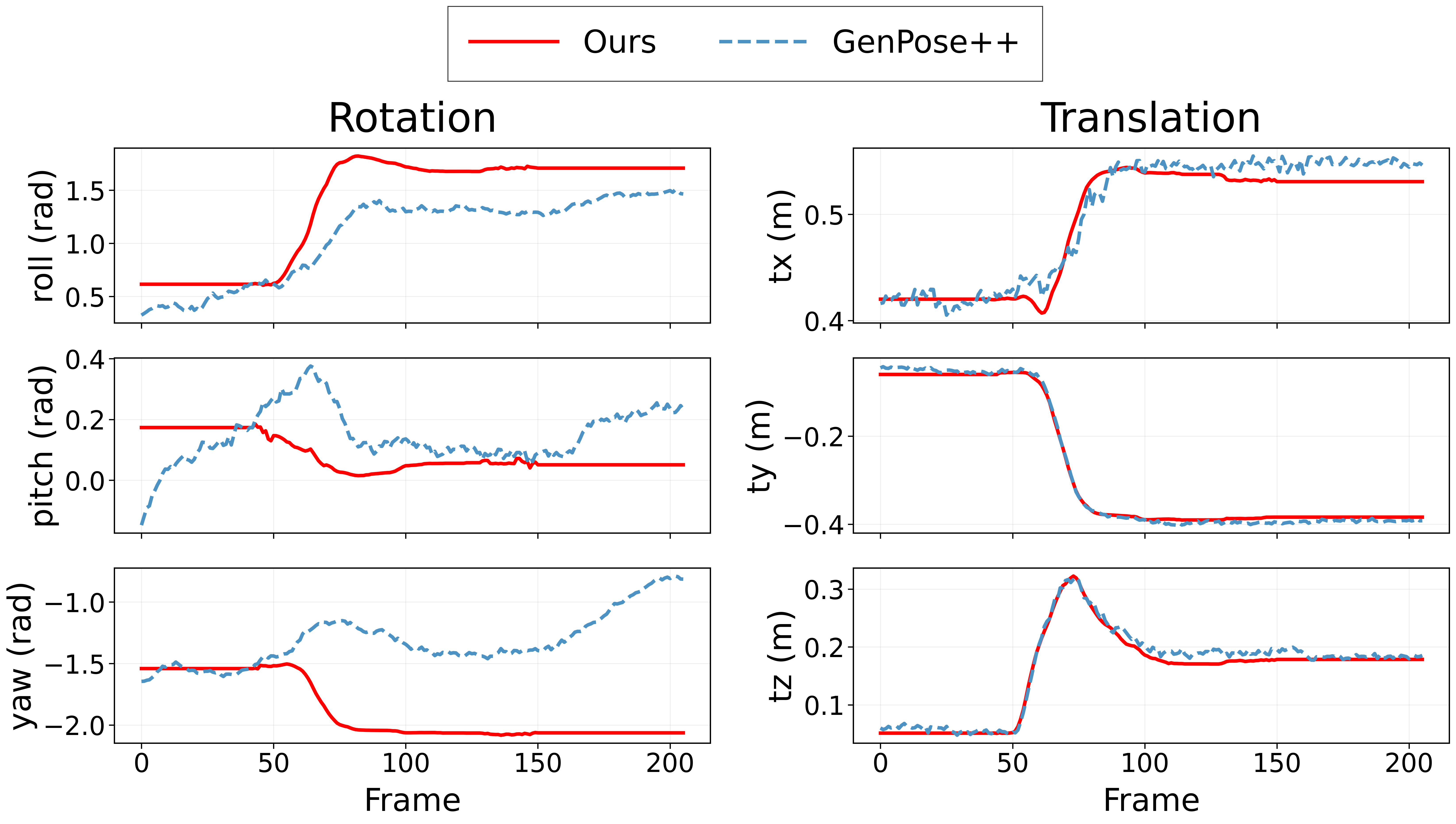}
    \caption{Qualitative Comparison of Pose Trajectories. The plots depict object rotation (left, in radians) and translation (right, in meters) across the sequence (Frame ID). Red solid lines represent our method, while blue dashed lines represent the GenPose++ baseline. Our framework achieves superior temporal stability while precisely capturing abrupt pose transitions during interaction events.}
    \label{fig:pose_curve}
\end{figure}
\paragraph{Pose Estimation Accuracy.}
We evaluate pose accuracy using RoboTwin ground-truth. As shown in Tab.~\ref{tab:pose_error}, our method achieves the lowest error, outperforming GenPose++ and HUGS. Specifically, we reduce $E_t$ to 0.5864 cm and $E_R$ to 5.185$^\circ$, a significant improvement over GenPose++ (0.9965 cm, 7.109$^\circ$). By leveraging kinematic data for refinement, our approach effectively mitigates estimation drift and ensures high-fidelity reconstruction.

\paragraph{Trajectory Analysis.}
Fig.~\ref{fig:pose_curve} visualizes real-world trajectories. Compared to the GenPose++ baseline, our method demonstrates superior temporal stability by suppressing high-frequency noise. Crucially, it accurately captures the abrupt pose transitions during gripper engagement (e.g., near Frame 50), which are typically obscured by noise in single-frame estimation methods. This ensures a stable trajectory during the motion phase while precisely reflecting discrete physical events during the skill phase.

\subsection{Data Augmentation}
\label{augmentation}

\begin{figure*}[!t]
    \centering
    \includegraphics[width=1\linewidth]{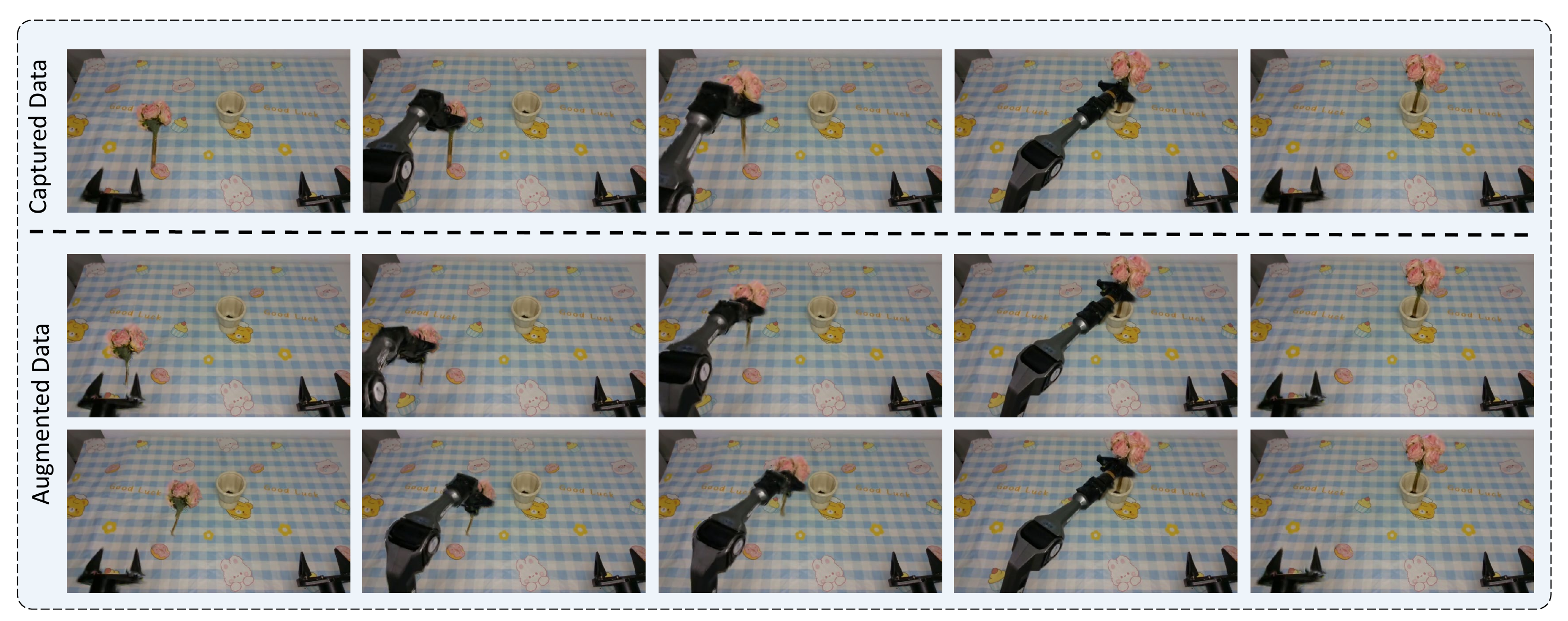}
    \caption{Data augmentation results. The first row shows the original grasping demonstration. The subsequent rows display high-fidelity augmentations where the flower and vase are spatially manipulated, demonstrating our method's ability to generate diverse training data from a single video.}
    \label{fig:augmentation}
\end{figure*}

Beyond reconstruction, our decoupled representation enables high-fidelity data augmentation. Figure~\ref{fig:augmentation} demonstrates this capability using a scene containing a flower and a vase. The first row displays the original grasping demonstration. The bottom two rows show augmented sequences where both the position and orientation of the flower are altered, while the vase remains stationary. Our method successfully generates photorealistic novel scenarios from a single source video, preserving interaction fidelity across different spatial configurations. This capability is particularly valuable for scaling datasets for downstream tasks, such as Vision-Language-Action (VLA) model training, where diverse object poses are critical for generalization.

\begin{table}[t]
  \centering
  \renewcommand{\arraystretch}{0.9}
  \begin{tabular*}{\columnwidth}{@{\extracolsep{\fill}}ccc|cc}
    \toprule
    $\mathcal{L}_{\text{pose}}$ & $\mathcal{L}_{\text{app}}$ & $\mathcal{L}_{\text{mask}}$ & PSNR~$\uparrow$ & SSIM~$\uparrow$ \\
    \midrule
    - & - & - & 30.08 & 0.932 \\
    \checkmark & - & - & 31.15 & 0.943 \\
    \checkmark & \checkmark & - & 32.24 & 0.949 \\
    \checkmark & \checkmark & \checkmark & \textbf{32.31} & \textbf{0.951} \\
    \bottomrule   
  \end{tabular*}
   \caption{Ablation study of $\mathcal{L}_{\text{pose}}$ (Pose constraint), $\mathcal{L}_{\text{app}}$ (Appearance alignment), and $\mathcal{L}_{\text{mask}}$ (Opacity mask supervision) on the RoboTwin dataset.}
  \label{tab:ablation}
\end{table}

We conduct an ablation study to evaluate the individual contributions of the pose constraint $\mathcal{L}_{\text{pose}}$, appearance alignment loss $\mathcal{L}_{\text{app}}$, and opacity mask supervision $\mathcal{L}_{\text{mask}}$. As shown in Tab.~\ref{tab:ablation}, $\mathcal{L}_{\text{pose}}$ provides the most significant performance gain, as it ensures geometric stability and reduces the high-frequency jitter inherent in sparse-view reconstruction. The introduction of $\mathcal{L}_{\text{app}}$ further refines the rendering quality by enforcing photometric consistency across the sequence. 

Notably, while the opacity-controlling $\mathcal{L}_{\text{mask}}$ yields a relatively marginal improvement in global metrics, it is indispensable for obtaining high-quality decoupled representations. By explicitly supervising the Gaussian opacity, $\mathcal{L}_{\text{mask}}$ effectively eliminates floaters and ensures sharp, physically plausible boundaries between the robot manipulator and the manipulated objects. This precise disentanglement is critical for downstream tasks like data augmentation, where clean object-environment separation is required for realistic re-composition.

%% file: sec/5_conclusion.tex
\section{Limitation and Future Work}
\label{sec:limit}
Despite its effectiveness in structured scene reconstruction, our framework still has limitations.
In particular, rendering quality may degrade when data augmentation introduces large spatial offsets from the original demonstration.
This is mainly due to the sparse-view nature of 3D Gaussian Splatting, where Gaussians optimized from limited viewpoints may lack sufficient geometric consistency for large-viewpoint generalization.

Several directions remain for future work.
First, replacing 3DGS with more geometrically constrained representations, such as 2D Gaussian Splatting (2DGS), could improve surface consistency and view synthesis quality.
Second, augmentation quality could be further enhanced by incorporating multiple reference demonstrations.
By fusing multi-state information from diverse interaction sequences of the same object, the framework could build a more robust scene representation and support higher-fidelity synthesis under large spatial perturbations.

\section{Conclusion}
\label{sec:con}
We present a unified framework for reconstructing interactive, simulation-ready Gaussian digital twins from real-world monocular egocentric robotic videos.
Our framework adopts a graph-based disentangled representation to separately model the robot, objects, and background, enabling object-level controllable reconstruction.
A task-oriented spatio-temporal alignment mechanism improves trajectory stability and rendering fidelity by separating free-space motion from contact-rich interaction phases.
To alleviate data sparsity, we further introduce a topology-preserving augmentation strategy that synthesizes physically consistent trajectories from a single demonstration.
Overall, our approach provides a scalable pipeline for converting raw monocular robotic videos into high-fidelity digital assets.